\documentclass[final,1p,times]{elsarticle}

\usepackage{amssymb}
\usepackage{amsmath}
\usepackage{booktabs}
\usepackage{svg}
\usepackage{url}
\svgsetup{inkscapelatex=false}
\newtheorem{remark}{Remark}
\graphicspath{{figures/}}
\newcommand{\BlackBox}{\rule{1.5ex}{1.5ex}}  
\ifdefined\proof
    \renewenvironment{proof}{\par\noindent{\bf \ }}{\hfill\BlackBox\\[2mm]}
\else
    \newenvironment{proof}{\par\noindent{\bf Proof\ }}{\hfill\BlackBox\\[2mm]}
\fi

\usepackage{environ}
\newcommand{\acksection}{\section*{Acknowledgments and Disclosure of Funding}}
\NewEnviron{ack}{%
  \acksection
  \BODY
}

\journal{Neural Networks}

\begin{document}

\begin{frontmatter}



\title{Variational Rank Reduction Autoencoders}


\author[PIMM,SKF]{Jad Mounayer\corref{cor1}} 
\cortext[cor1]{Corresponding author}
\ead{jad.mounayer@outlook.com}

\author[Sara]{Alicia Tierz}
\author[SKF]{Jerome Tomezyk}
\author[UNF]{Chady Ghnatios}
\author[PIMM,CNRS]{Francisco Chinesta}

\affiliation[PIMM]{organization={PIMM Laboratory, ENSAM},
            addressline={Blvd de l'Hôpital}, 
            city={Paris},
            postcode={75013}, 
            country={France}}

\affiliation[Sara]{organization={I3A, Universidad de Zaragoza},
            addressline={C. María de Luna}, 
            city={Zaragoza},
            postcode={50018}, 
            country={Spain}}

\affiliation[SKF]{organization={SKF Magnetic Mechatronics},
            addressline={Rue des Champs}, 
            city={Saint-Marcel},
            postcode={27950}, 
            country={France}}

\affiliation[UNF]{organization={University of North Florida},
            addressline={1 UNF Dr.}, 
            city={Jacksonville},
            postcode={32224}, 
            country={United States}}

\affiliation[CNRS]{organization={CNRS@Create},
            addressline={1 CREATE Way, CREATE Tower}, 
            city={Singapore},
            postcode={138602}, 
            country={Singapore}}

\begin{abstract}
Deterministic Rank Reduction Autoencoders (RRAEs) enforce by construction a regularization on the latent space by applying a truncated SVD. While this regularization makes Autoencoders more powerful, using them for generative purposes is counter-intuitive due to their deterministic nature. On the other hand, Variational Autoencoders (VAEs) are well known for their generative abilities by learning a probabilistic latent space. In this paper, we present Variational Rank Reduction Autoencoders (VRRAEs), a model that leverages the advantages of both RRAEs and VAEs. Our claims and results show that when carefully sampling the latent space of RRAEs and further regularizing with the Kullback-Leibler (KL) divergence (similarly to VAEs), VRRAEs outperform RRAEs and VAEs. Additionally, we show that the regularization induced by the SVD not only makes VRRAEs better generators than VAEs, but also reduces the possibility of posterior collapse. Our results include a synthetic dataset of a small size that showcases the robustness of VRRAEs against collapse, and three real-world datasets; the MNIST, CelebA, and CIFAR-10, over which VRRAEs are shown to outperform both VAEs and RRAEs on many random generation and interpolation tasks based on the FID score. We developed an open-source implementation of VRRAEs in JAX \cite{jax} (Equinox\cite{eqx}), available at \url{https://github.com/JadM133/RRAEs.git}.
\end{abstract}




\end{frontmatter}



\newpage
\section{Introduction}
Many real-life processes are probabilistic. Researchers, for decades, have been highlighting the variational aspect of the world \cite{planck1900zur, wetterich2020probabilistic}. Consequently, if we are to reproduce a certain process, probabilistic models are appealing. When learning to reproduce the probability distribution of a process, we don't only learn a simple ``input/output'' relationship; we are also capable of replacing the process with the model, thus generating new samples. 

More precisely, let $X$ be a set of training samples, following a certain distribution $a(X)$. Let us assume we have a model that can generate new samples with a distribution $b(X)$. We aim to encourage $a(X)$ and $b(X)$ to be as close as possible. Several techniques have been previously used for this type of generative modeling, but with various limitations. For instance, Gaussian Mixture Models (GMMs) \cite{reynolds2009gaussian, mcnicholas2008parsimonious} made strong assumptions about the structure of the data. Other techniques, such as Bayesian Networks \cite{pearl1988probabilistic, govan2018bayesiannetwork, govan2023bayesiannetwork} or Hidden Markov Models \cite{rabiner1986introduction, harte2021hiddenmarkov, bourlard1994hidden}, require an extensive inference procedure based on Markov Chain Monte Carlo (MCMC) \cite{gelman1996markov, tierney1994markov}.

On the other hand, as Neural Networks gained popularity due to their nonlinear capacities and speed during evaluation, Variational Autoencoders (VAEs) \cite{kingma2013autoencoding} were introduced to learn probabilistic models while avoiding the limitations mentioned above. The goal of VAEs is to have a model that uses latent variables to sample the data. VAEs maximize the log likelihood of generating the training data by minimizing two objective functions,
\begin{equation*}
    \mathcal{L}_{VAE} = \mathcal{L}_{rec} + \mathcal{L}_{KL}.
\end{equation*}
The first term maximizes the probability of the training data conditional on the latent variables, while the second term (called Kullback-Leibler (KL) divergence) enforces the predicted latent distribution to be as close as possible to a chosen distribution (usually standard normal), called the true prior \cite{doersch2021tutorialvariationalautoencoders, DBLP:journals/corr/abs-1906-02691}.

The generative abilities of VAEs have made them very popular among probabilistic models. They have been used in many applications, including in the medical \cite{laptev2021generative}, chemical \cite{lovric2021should}, and engineering \cite{yan2020chiller, regenwetter2022deep} fields. VAEs are also the basic construction block for more complex models. For instance, some papers included discriminators \cite{gao2020zero, niu2020lstm}, others used regularization techniques \cite{wu2020vector, hadjeres2017glsr}, or replaced the KL divergence with an other regularizing term \cite{tolstikhin2019wassersteinautoencoders}. While VAEs inherit the approximation abilities of Neural Networks and are fast during inference, they face two main limitations:

\begin{enumerate}
    \item The KL divergence is the only regularization applied in a VAE. Accordingly, minimizing this term in some cases becomes crucial to obtain a meaningful latent space. Since during optimization we are trying to enforce both the reconstruction quality and the KL divergence, VAEs usually end up with worse reconstructions (i.e., blurrier images, in the case of image outputs) \cite{bredell2023explicitlyminimizingblurerror, dalal2024shorttimefouriertransformdeblurring, khan2018adversarialtrainingvariationalautoencoders}. 
    \item Posterior collapse \cite{he2019lagginginferencenetworksposterior, havrylov2020preventingposteriorcollapselevenshtein, dang2024vanillavariationalautoencodersdetecting}, which happens especially when the decoder is too large for the given amount of data. In this case, the probability distribution predicted by the encoder collapses to the true prior, thus deteriorating the generative abilities of VAEs.
\end{enumerate}

On the other hand, recently, Rank Reduction Autoencoders (RRAEs) \cite{mounayer2025rankreductionautoencoders} have been presented as a more stable version of Vanilla AEs. The main idea behind RRAEs is to include a truncated SVD in the latent space during training, enforcing the bottleneck by reducing the rank of the latent matrix instead of its dimension. Despite their deterministic nature, RRAEs have been shown to outperform other regularizing Autoencoders on tasks such as interpolation and random generation in their latent space. This is mainly due to the regularizing characteristics of the truncated SVD, without adding any terms to optimize in the loss. Yet, as RRAEs are not probabilistic models, they do not learn a distribution and therefore are less suitable to be used as generative models.

In this paper, we introduce Variational Rank Reduction Autoencoders (VRRAEs), a probabilistic version of RRAEs that learns a generative distribution of the data while benefiting from the regularization of the truncated SVD. We show throughout the paper, through conceptual and empirical evidence, that VRRAEs help mitigate both VAE's limitations mentioned above (i.e., further regularization and robustness to posterior collapse). The resulting architecture produces clearer images and outperforms RRAEs and VAEs on MNIST, CelebA, and CIFAR-10 on almost all interpolation and random generation tasks. Furthermore, we revisit the dataset proposed by \cite{mounayer2025rankreductionautoencoders}, but with a much smaller size (100 training images in total). We show that, on such a small dataset with local behavior, a typical VAE struggles to generate new samples due to posterior collapse, while VRRAEs overcome this limitation.

\section{Background}
We begin by defining the necessary notation and use those to present VAEs and RRAEs, both crucial for defining our model later.

Without loss of generality, let $X\in\mathbb{R}^{D\times N}$ be a set of $N$ data samples, each of dimension $D$ (inputs of higher dimensions, such as images, can be flattened). Let $X_j$ be the $j$-th sample of $X$ (or $j$-th column), and $k^*$ be the bottleneck size down to which we want to compress our data.

\begin{figure}
    \centering
    \includegraphics[width=1\linewidth, trim=0 0 0 0cm, clip]{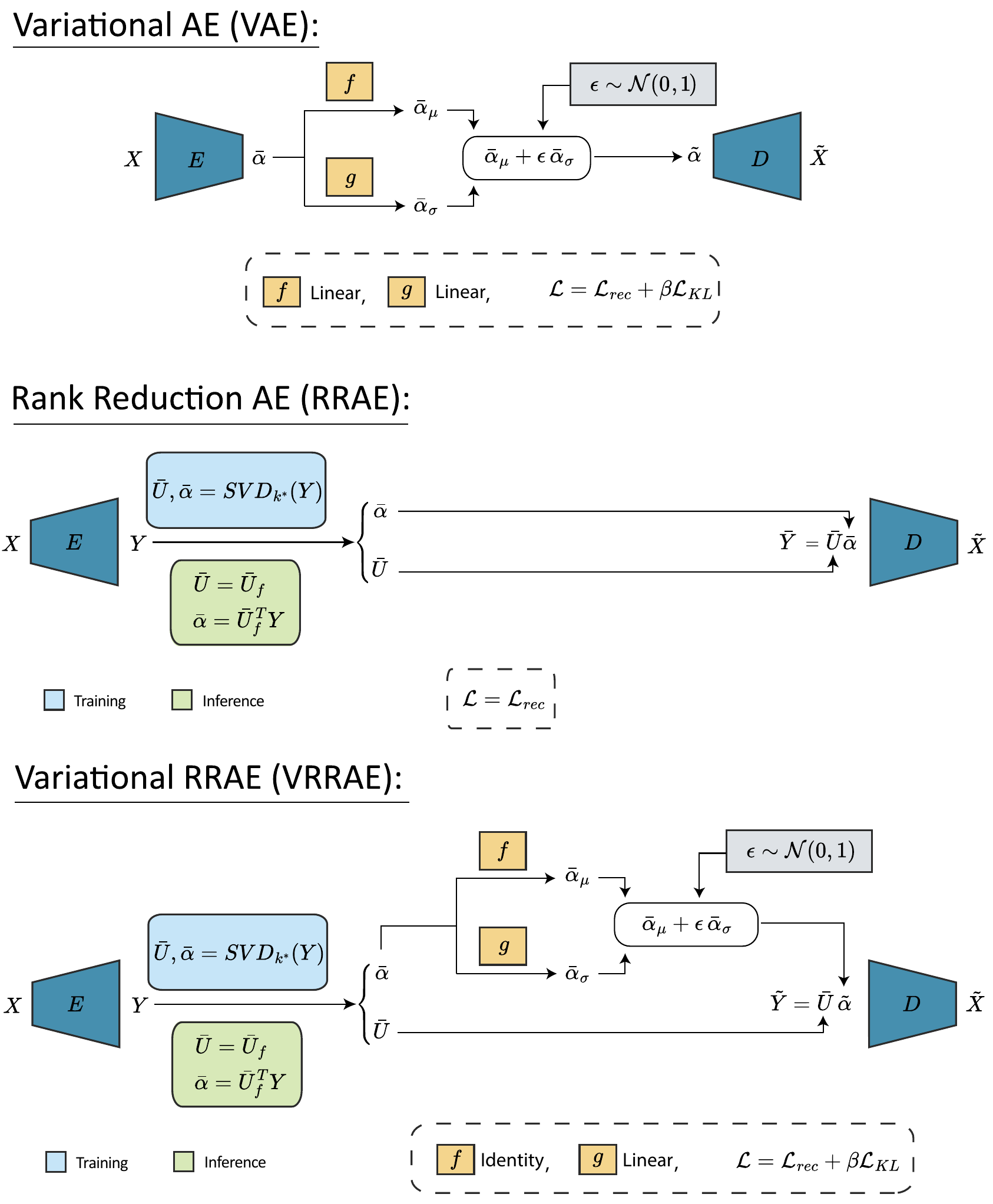}
    \caption{Schematic illustrating the architecture of Variational Rank Reduction Autoencoders (VRRAEs). Both $E$ and $D$ are trainable Neural Networks representing an encoding and a decoding map. $SVD_{k^*}$ is a truncated SVD of rank $k^*$ as detailed in Equation \ref{eq:RRAE}. Note that $f=I$ (identity) for VRRAEs.}
    \label{fig:Network_structure}
\end{figure}

\underline{Variational Autoencoder (VAE):} A Gaussian VAE, illustrated in the top part of Figure \ref{fig:Network_structure}, can be defined as follows,
\begin{equation*}
\begin{cases}
    \bar{\alpha} = E(X), \qquad \begin{cases}
            \bar{\alpha}_{\mu} = f(\bar{\alpha}) \\[2ex] \bar{\alpha}_{\sigma} = g(\bar{\alpha})
        \end{cases}, \qquad \tilde{\alpha} = \bar{\alpha}_{\mu} +\epsilon\, \bar{\alpha}_{\sigma}, \qquad \tilde{X}=D(\tilde{\alpha}), \\[5ex]
        \text{with,}  \qquad f: \mathbb{R}^{k^*\times N} \xrightarrow{} \mathbb{R}^{k^*\times N} \quad \text{Linear}, \,\,\quad g: \mathbb{R}^{k^*\times N} \xrightarrow{} \mathbb{R}^{k^*\times N} \quad \text{Linear}, \\[2ex]
        \text{and,} \, \, \qquad\epsilon\sim \mathcal{N}(0, I),  \qquad  E: \mathbb{R}^{D\times N}\xrightarrow{}\mathbb{R}^{k^*\times N}, \qquad D: \mathbb{R}^{k^*\times N}\xrightarrow{}\mathbb{R}^{D\times N}.
\end{cases}
\end{equation*}
In the above, the latent space $\tilde{\alpha}$ is modeled as being sampled from a conditional distribution where $q(\tilde{\alpha}\mid X_j) = \mathcal{N}((\bar{\alpha}_\mu)_j, (\bar{\alpha}_\sigma)_j)$, for all $j\in[1, N]$, and the decoder models $p(X_j\mid \tilde{\alpha}_j)$ as a distribution centered at the deterministic output $\tilde{X}_j$.

The objective of VAEs is to maximize the likelihood of the training data, which can be done by minimizing the following loss function \cite{kingma2013autoencoding}, 
\begin{equation}\label{eq:VAE_loss_p}
    \mathcal{L}_{VAE} = \frac{1}{N}\sum_{j=1}^N\left(
    \, \mathbb{E}_{q(\tilde{\alpha} \mid X_j)} \left[ -\log p(X_j \mid \tilde{\alpha}) \right] \,
    + \, \mathrm{KL}\left(q(\tilde{\alpha} \mid X_j) \,\|\, p(\tilde{\alpha})\right)\,\right),
\end{equation}
where $p(\tilde{\alpha})$ is a true prior, to which we want our latent distribution to resemble. When the prior is chosen to be standard normal with $p(\tilde{\alpha})=\mathcal{N}(0, I)$, and the decoder outputs a deterministic output $\tilde{X}$, the loss in Equation \eqref{eq:VAE_loss_p} can be re-written as \cite{kingma2013autoencoding},
\begin{equation}\label{eqn:loss_vae}
    \mathcal{L}_{VAE}=\underbrace{\|X-\tilde{X}\|_2}_{\mathcal{L}_{rec}} + \beta\underbrace{\frac{0.5}{N}\,\,\text{sum}(\mathbf{1}_{k^*\times N}+\log(\bar{\alpha}_\sigma\odot\bar{\alpha}_\sigma)-\bar{\alpha}_\mu\odot \bar{\alpha}_\mu-\bar{\alpha}_\sigma\odot\bar{\alpha}_\sigma)}_{\mathcal{L}_{KL}},
\end{equation}
where $\mathbf{1}_{k^*\times N}$ is a matrix of ones of dimension $(k^*\times N)$, $\odot$ is the Hadamard product (or element-wise multiplication), $\beta$ is a constant that scales the variance of the output distribution \cite{doersch2021tutorialvariationalautoencoders}, $\|\cdot\|_2$ is the $L2$ norm, and ``sum'' indicates the sum of all of the entries of a matrix. Note that we choose to use the $L2$ norm for the reconstruction error throughout the paper, even though any other norm could have been used as well.

The main advantage of VAEs is that they learn a probabilistic model that's able to generate new samples once training is done. The second term of the loss (or the KL divergence) encourages the sample distributions to be closer to each others, leading to a more relevant latent space. 

On the other hand, Variational Autoencoders present two main limitations. First, since the KL divergence is the only regularization, and the loss to optimize combines both the reconstruction error and the KL divergence, the reconstruction error in some cases has to remain high if the space is to be regularized enough. Second, if the decoder is too large with respect to the complexity or size of the dataset, the training may steer the encoder to satisfy $q(\tilde{\alpha}|X_j) \approx p(\tilde{\alpha})$, a phenomenon known as posterior collapse. While this reduces the magnitude of the KL loss term, it also causes the latent representation to lose its dependence on the input $X_j$, leading to an uninformative latent space and an increased reconstruction error.

\underline{Rank Reduction Autoencoder (RRAE):} Using the same notation as VAEs, and for a chosen latent space size $L$, RRAEs, illustrated in the middle of Figure \ref{fig:Network_structure}, can be defined as follows,
\begin{equation}\label{eq:RRAE}
\begin{cases}
    Y = E(X) = US V^T = U\alpha, \qquad \displaystyle \bar{Y}=\sum_{i=1}^{k^*}U_is_iV_i^T=\bar{U}\bar{S}\bar{V}^T=\bar{U}\bar{\alpha}, \qquad \tilde{X} = D\left(\bar{Y}\right), \\[4ex]
    \text{where,} \quad  E: \mathbb{R}^{D\times N}\xrightarrow{}\mathbb{R}^{L\times N}, \quad\bar{Y}\in\mathbb{R}^{L\times N},\quad\bar{\alpha}\in\mathbb{R}^{k^*\times N},\quad\quad D: \mathbb{R}^{L\times N}\xrightarrow{}\mathbb{R}^{D\times N},
\end{cases}
\end{equation}
where $USV^T$ represents the singular value decomposition of a matrix, and $\left(\bar{\cdot}\right)$ represents a matrix after the truncated SVD (e.g., $\bar{Y}$ is the latent space after truncation, and $\bar{\alpha}$ are the truncated SVD coefficients). 

Note that during inference, the SVD is replaced by the projection over the basis found during training $\bar{U}_f$ (refer to  \cite{mounayer2025rankreductionautoencoders} for more details). In RRAEs, the bottleneck is not in the dimension of the latent space $\bar{Y}$, but in the rank of the truncated latent matrix, illustrated by the coefficients $\bar{\alpha}=\bar{S}\bar{V}^T$. Accordingly, both encoding/decoding maps $E$ and $D$ do not depend on the choice of the bottleneck $k^*$. Since the bottleneck is represented by $\bar{\alpha}$, it inherits some of the properties of the truncated singular values $\bar{S}$ and the truncated right singular vectors $\bar{V}^T$, mainly:
\begin{enumerate}
    \item The regularization of the bottleneck. For every batch of size $B$, and latent dimension $i$, the bottleneck satisfies $\sum_{j=1}^{B}\bar{\alpha}_{i,j}^2=\bar{s}_i^2$, since the right singular vectors are orthonormal.
    \item The bottleneck is sorted by decreasing importance, since the singular values $s_i$ are sorted from the largest to the smallest.
\end{enumerate}
Note that since the regularization is imposed strongly inside the network, the loss only consists of one term,
\begin{equation*}
    \mathcal{L}_{RRAE}=\mathcal{L}_{rec}=\|X-\tilde{X}\|_2.
\end{equation*}
The regularization in RRAEs has been shown to improve the behavior of autoencoders in \cite{mounayer2025rankreductionautoencoders}. However, RRAEs are deterministic Autoencoders, which makes them less suitable for generating new samples in the latent space.

\section{Proposed Model}\label{sec:methodology}
In this part, we present the core of the paper, Variational Rank Reduction Autoencoders (VRRAEs), an architecture designed to benefit from the advantages of both RRAEs and VAEs. Note that since the bottleneck in RRAEs is represented by $\bar{\alpha}$, only the truncated coefficients $\bar{\alpha}$ are sampled instead of sampling the reconstructed latent space $\bar{Y}$. As can be seen in the bottom of Figure \ref{fig:Network_structure}, the SVD coefficients are sampled, similarly to how the bottleneck is sampled in VAEs (with a main difference discussed just after in the remark). The resulting, probabilistic, SVD coefficients $\tilde{\alpha}$ are then multiplied by the deterministic basis previously found $\bar{U}$ before being passed to the decoder. 

\begin{remark}
    A main difference between VRRAEs and VAEs is that the means are found using $f=I$, the identity map. In other words, the expected values values $\mathbb{E}(\tilde{\alpha}_{i,j})=\bar{\alpha}_{i,j}$, $\forall i,j$, are the coefficients found by the SVD of $Y$. As detailed in \ref{ap:naive}, and shown empirically in the ablation study in Section \ref{sec:ablation}, this is crucial to preserve the properties and advantages of RRAEs.
\end{remark} 

The VRRAE architecture presented above tackles both VAE's limitations presented in the introduction as follows,

\begin{enumerate}
    \item \underline{Further Regularization (enforced strongly).}
    Note that one of the main advantages of RRAEs is the regularization imposed on the bottleneck. By choosing $f=I$, the sampled SVD coefficients retain the regularizing property. In other words, for any batch $b$ of size $B$, we can say that, $\mathbb{E}\left(\sum_{j=1}^B(\tilde{\alpha}^b_{i,j})^2\right)=\bar{s}_i$,  $\forall i\in[1, k^*]$, with $\tilde{\alpha}^b_{i,j}$ being the coefficients of batch $b$. Since this is enforced in a strong manner in the network, without adding any objectives to the optimization process, the quality of the generated samples is enhanced without causing an increase in the reconstruction error. We show empirically throughout the paper that this regularization helps in generating sharper images and achieving better FID scores. 
    
    \item \underline{Posterior Collapse: Less likely in VRRAEs.}
    Posterior collapse occurs when the encoder learns to predict the true prior, independently of the data sample given as input. In other words, the bottleneck's dependence on the input samples becomes minimal. Since our samples are stacked as columns in $X$, we can write posterior collapse of a certain dimension $i$ more formally as,
    \begin{equation}\label{eqn:zeta}
        \mathbb{E}(\tilde{\alpha}_{i,j}) \approx\zeta_i, \qquad \forall j\in[1, N],
    \end{equation}
    where $\zeta_i$ is the value to which the mean of dimension $i$ will collapse in the latent space. This phenomenon is likely to happen in VAEs since the nonlinearities could converge to any value $\zeta_i$, for any dimension $i$. However, for Equation \eqref{eqn:zeta} to hold and posterior collapse to happen in VRRAEs, we would need, for all $j\in[1, N]$, the following to be true.
    \begin{equation}\label{eqn:jesaispas}
        \mathbb{E}(\tilde{\alpha}_{i,j}) = \bar{\alpha}_{i,j} \approx\zeta_i \qquad \xrightarrow{}\qquad \bar{s}_i\bar{V}_{i,j}^T \approx\zeta_i \qquad \xrightarrow{}\qquad \bar{V}_{i,j}^T \approx \frac{\zeta_i}{\bar{s}_i}.
    \end{equation}
    Yet, we recall that $\bar{V}^T$ is orthonormal, therefore, and $\forall i$,
    \begin{equation}\label{eqn:final_zeta}
        \sum_{j=1}^N\left(\bar{V}^T_{i,j}\right)^2=1 \quad \xrightarrow[\text{Equation \eqref{eqn:jesaispas}}]{}\quad  \sum_{j=1}^N \frac{\zeta_i^2}{\bar{s}_i^2} \approx1 \quad \xrightarrow[\text{No dependence on $j$}]{} \quad \zeta_i \approx \frac{\pm \bar{s}_i}{\sqrt{N}}.
    \end{equation}
    The result in Equation \eqref{eqn:final_zeta} shows that the means in VRRAEs can only collapse to specific values of $\zeta_i$, as opposed to VAEs which can collapse to any value. This regularizes VRRAEs rendering them more robust to posterior collapse, as shown in Section \ref{sec:synthetic} when training both VAEs and VRRAEs on a dataset with local behavior and only 100 training samples.
\end{enumerate}
Further, note the VRRAEs, similarly to VAEs, regularize their latent space with the KL divergence. In practice, the same loss is implemented for both VAE and VRRAEs. Yet, note that when choosing the true prior to be $p(\tilde{\alpha}) = \mathcal{N}(0, I)$, and by leveraging the properties of the SVD coefficients and the fact that $f=I$, the KL divergence for VRRAEs can be expressed as\footnote{A detailed proof can be found in \ref{ap:KL}.},
\begin{equation}\label{eq:final_eq_kl_vrrae}
    \sum_{j=1}^N\mathrm{KL}\left(q(\tilde{\alpha} \mid X_j) \,\|\, p(\tilde{\alpha})\right) = 0.5\,\,\text{sum}(\mathbf{1}_{k^*\times N}+\log(\bar{\alpha}_\sigma\odot\bar{\alpha}_\sigma)-\left(\text{diag}(\bar{S})\right)^2-\bar{\alpha}_\sigma\odot\bar{\alpha}_\sigma),
\end{equation}
where we used the same notation as Equation \eqref{eqn:loss_vae}, and $\bar{S}$ is the truncated matrix containing the singular values on its diagonal, as defined in Equation \eqref{eq:RRAE}. The term representing the mean, $(\text{diag}(\bar{S}))^2$, does not depend on the sample $j$, and simply consists of the square of the singular values. In other words, and as shown empirically in \ref{ap:KL}, VRRAEs not only enforce a low rank of the latent matrix, but also encourage the singular values to be bounded. Though, just like VAEs, one has to find a compromise between bounding the singular values (minimizing the KL divergence), and reducing the reconstruction error (further details about the process of choosing $\beta$ for both VAEs, and VRRAEs can be found in \ref{ap:model}).

\section{Results}
This section presents empirical results to support the claims made when presenting VRRAEs. Throughout, we compare VRRAEs to three other models: a baseline diabolo AE, a Rank Reduction Autoencoder, and a Variational Autoencoder. VRRAEs are also compared to other deterministic Autoencoders in \ref{ap:other}. We do not compare our model with extensions of VAEs (e.g., discriminators, annealing of $\beta$ values, or using other regularizers than the KL-divergence), as these changes could be similarly applied to VRRAEs to improve their performance. 

For training, all AE models share the same base architecture (i.e., encoder and decoder), which is detailed in \ref{ap:model}. The main differences between the models are the size of the latent space, set to $k^*$ for the Diabolo AE and the VAE, and to $L$ for the RRAE and the VRRAE (with a bottleneck of $k^*$ enforced via truncated SVD). Further, the optimal values of $\beta$ scaling the KL divergence are found by testing multiple values for both VAEs and VRRAEs and choosing the best model. The chosen values of $k^*$, $L$, and the process of choosing $\beta$ are given in \ref{ap:model}.

Readers interested in more details about the training times, or the effect of the batch size can refer to both \ref{ap:time}, and \ref{ap:bs}. Further, the code used to produce the results in this paper is part of an open-source library in JAX \cite{jax} (Equinox \cite{eqx}), available at \url{https://github.com/JadM133/RRAEs.git}.

\begin{table}[!b]
  \caption{Relative errors (in \%) when training different models using 5 different seeds over the shifted Gaussian dataset with 100 training curves. The error for random generation is computed by comparing the generated curve with a fitted Gaussian of the right magnitude and spread.}
  \label{sample-table}
  \centering
  \begin{tabular}{ccc}
    \toprule
    Model & Test Error & Random Gen. Error \\
    \midrule
    Diabolo & $23.24 \pm 11.26$ &   $21.28 \pm 11.32$   \\
    VAE & $26.31 \pm 22.07$ & $9.47 \pm 5.76$  \\
    RRAE & $56.05 \pm 30.21$&  $40.58 \pm 18.51$  \\
    VRRAE & $\mathbf{10.03 \pm 8.96}$ & $\mathbf{5.88 \pm 2.94}$ \\
    \bottomrule
\end{tabular}
\end{table}

\begin{figure}[!b]
  \centering
  \includegraphics[width=1\textwidth]{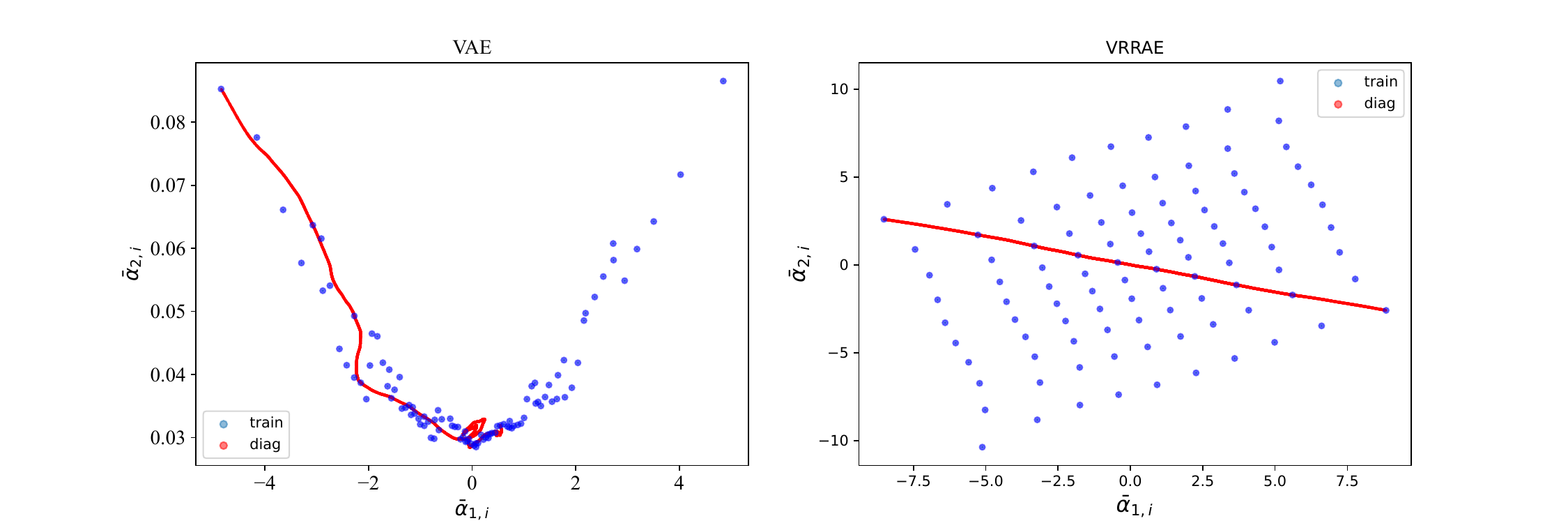}
  \caption{Visualization of the second latent mean plotted against the first one for a certain seed in the latent space for VAEs (left) and VRRAEs (right). The collapse of the second latent dimension in VAEs is evidenced by the small variations along the $y$-axis (i.e., $0.02\leq\zeta_i\leq0.09$ in Equation \eqref{eqn:zeta}). In contrast, VRRAEs do not suffer from collapse, and preserve the structure of a diagonal motion (highlighted in red), maintaining interpretability.}
  \label{fig:posteriors_gauss}
\end{figure}

\subsection{Synthetic Dataset} \label{sec:synthetic}
In this section, we test the Autoencoders discussed above on a challenging synthetic dataset to showcase the robustness of VRRAEs to posterior collapse. We chose the same dataset presented in \cite{mounayer2025rankreductionautoencoders}, which consists of Gaussian bumps with a fixed spread and magnitude, on a 2D grid, that are moved along both the $x$ and $y$ axes. However, we reduced the number of training samples from $600$ to only $100$ (i.e., $10$ grid points in both $x$ and $y$). The small size of the dataset, as well as the challenging local behavior, makes it harder for autoencoders to find meaningful latent spaces without either collapsing the posterior or overfitting. To test the robustness of the AE architectures, we trained all four models 5 times with different seeds for initializing the network. We then randomly generated samples from the latent space of each Autoencoder by using a Gaussian Mixture Model. We quantified how well the generated images were by fitting a Gaussian to each new sample and finding the relative difference between the image and the fitted Gaussian with the right magnitude. The mean and standard deviation of the relative errors for random generation, as well as the reconstruction errors over a test set of $10000$ randomly chosen Gaussian curves can be found in Table \ref{sample-table}.

The large standard deviation for VAEs shows that they might collapse their posterior depending on the seed. To further investigate wether it is posterior collapse, we visualize the expected value of the predicted posterior of each sample for a selected seed, in both VAEs and VRRAEs, by scattering them on a 2D plot in Figure \ref{fig:posteriors_gauss}. The small values of the means on the $y$-axis for VAEs shows that the second dimension of the latent space collapsed. On the other hand, we plot in red the latent space for many test gaussian images that represent a diagonal motion in the real space. As can be seen in the latent space of VRRAEs (Figure \ref{fig:posteriors_gauss}), the diagonal motion remains a diagonal motion in the latent space, which shows the interpretability of the latent space of VRRAEs.

\begin{figure}[!t]
  \centering
  \includegraphics[width=0.8\textwidth]{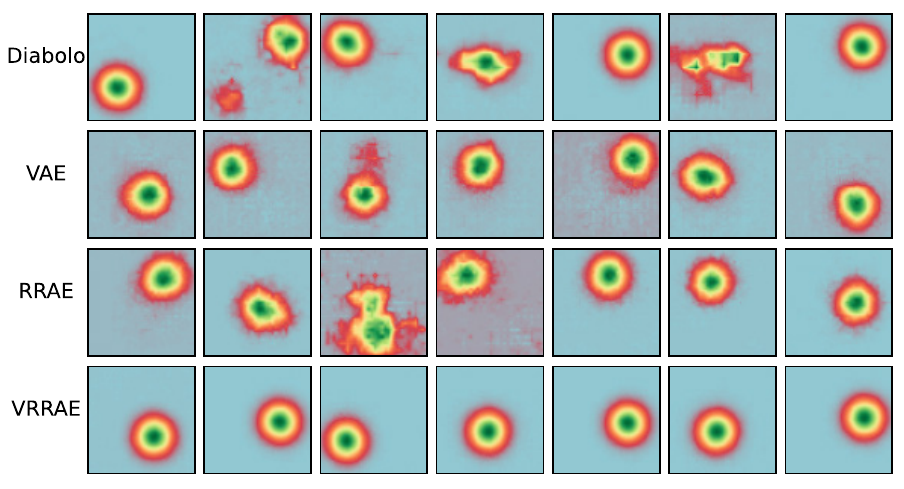}
  \caption{Randomly generated samples for different architectures on the 2D gaussian problem}
  \label{fig:random_gauss}
\end{figure}

A few randomly generated samples can be found in Figure \ref{fig:random_gauss}. The results showcase the stability of VRRAEs, caused by the regularization of the truncated SVD as claimed in the previous section.

\subsection{Real World data}\label{sec:real-world}

We compare VRRAEs to the other proposed models on three real-world datasets: the MNIST, CIFAR-10, and CelebA. To evaluate their performance, autoencoders' test reconstruction error, as well as the Fréchet Inception Distance (FID\footnote{More details about how the FID was computed can be found in \ref{ap:FID}.}) score for both interpolation and random generation, are reported. For interpolation, we randomly sample $250$ couples of images and interpolate linearly in the latent space to get 5 new generated images on which the FID score is computed. Random generation is performed by using a Gaussian Mixture Model on the latent space and randomly sampling 1250 images from it. The results, documented in Table \ref{tbl:rw_results}, show that VRRAEs are able to achieve a better performance than both RRAEs and VAEs on almost all the datasets presented. First, we note that the reconstruction error on the test set is always smaller for VRRAEs. This is mainly due to the fact that the KL-divergence isn't as prominent as in VAEs since there is an existing regularization in RRAEs. The lower reconstruction error, together with the regularization from both the KL divergence and the truncated SVD, allows VRRAEs to achieve the best FID scores on almost all the datasets. More details about the hyperparameters and how fair comparative training was achieved for all autoencoders can be found in \ref{ap:model}.

\begin{table}[!t]
  \caption{Quantitative comparison across datasets. The third column for each dataset represents the test reconstruction error, while the first two columns document the mean FID score (over 5 random seeds) for interpolation and random generation respectively. Standard deviations have been omitted since they're of small magnitudes.}
  \label{tbl:rw_results}
  \centering
  \begin{tabular}{cccccccccc}
    \toprule
    {} & \multicolumn{3}{c}{MNIST} & \multicolumn{3}{c}{CIFAR-10} & \multicolumn{3}{c}{CelebA} \\
    \cmidrule(lr){2-4} \cmidrule(lr){5-7} \cmidrule(lr){8-10}
    Model & Inter. & Rand. & Rec. & Inter. & Rand. & Rec. & Inter. & Rand. & Rec. \\
    \midrule
    AE     & 7.31  & 40.12 & 27.31  & 143.31 & 140.35 & 18.39 & 15.24 & 15.88 & 18.82 \\
    VAE    & 11.95  & \textbf{30.63} & 32.602  & 137.74 & 135.77 & 17.86 & 8.94 & 9.66 & 16.55 \\
    RRAE   & 6.68  & 45.46 & 27.90  & 140.91 & 136.94 & 17.99 & 13.27 & 13.70 & 17.52 \\
    \textbf{VRRAE}  & \textbf{5.89}  & 38.77 & \textbf{26.00}  & \textbf{129.68} & \textbf{129.89} & \textbf{17.04} & \textbf{7.06}  & \textbf{7.60}  & \textbf{15.03} \\
    \bottomrule
  \end{tabular}
\end{table}

\begin{figure}[!ht]
  \centering
  \includesvg[width=0.8\textwidth]{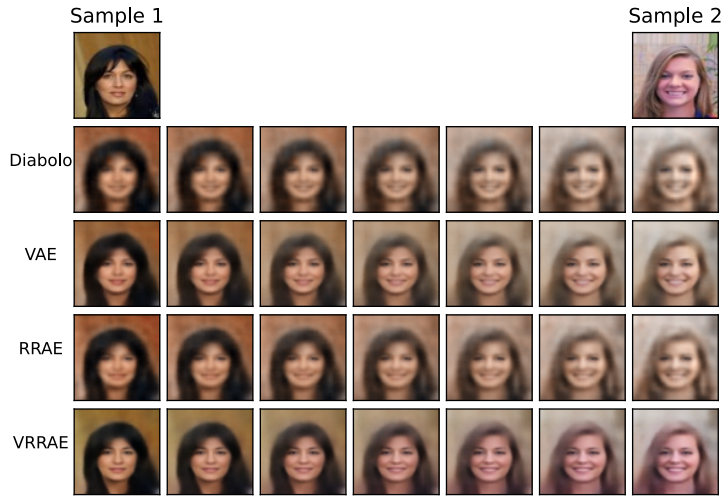}
  \vspace{-0.8cm}
  \caption{Example of an interpolation (linear, in the latent space) between two CelebA samples. Note that the VRRAE is the only model to capture the right skin color, even though all models have the same bottleneck size.}
  \label{fig:interp_celeba}
\end{figure}

\begin{figure}[!ht]
  \centering
  \includesvg[width=0.75\textwidth]{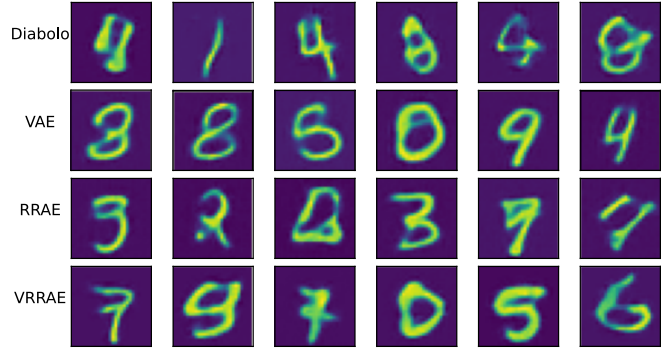}
  \caption{Randomly generated MNIST samples for each one of the selected models.}
  \label{fig:random_mnist}
\end{figure}

\begin{figure}[!ht]
  \centering
  \hspace{-0.5cm}
  \includesvg[width=0.65\textwidth]{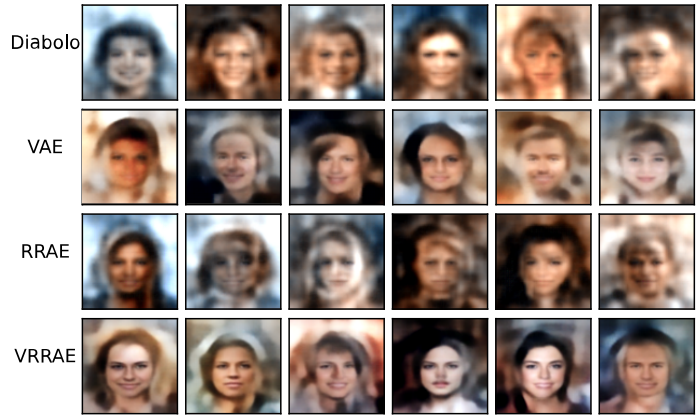}
  \caption{Randomly generated CelebA samples with each one of the selected Autoencoders.}
  \label{fig:random_celeba}
\end{figure}

We present some examples of interpolated images on the CelebA dataset in Figure \ref{fig:interp_celeba}. We also show some generated MNIST and CelebA samples in Figures \ref{fig:random_mnist}, and \ref{fig:random_celeba}. More interpolated/generated images of all three datasets can be found in \ref{ap:cifar-samples}. Overall the results show that VRRAEs outperform both deterministic RRAEs and VAEs, for the exception of the MNIST random generation where the SVD regularization doesn't seem to be as important.

\section{Ablation Study}\label{sec:ablation}
In this section, we investigate empirically wether the theory behind choosing $f=I$ checks out in practice. We also study the effect of the KL divergence term in the loss and show that it is required.

Therefore, we compare VRRAEs on all datasets with three architectures,
\begin{enumerate}
    \item \underline{RRAE + VAE:} A naive combination between an RRAE and a VAE, with $f$ being a trainable linear map instead of the identity. More details behind why this architecture is expected to perform worse can be found in \ref{ap:naive}.
    \item \underline{VAE (f=I):} A typical variational autoencoder but with the identity as its map for the mean in the latent space. This is mainly to show that the good performance of VRRAEs is not due to $f=I$ being a suitable map in the latent space.
    \item  \underline{VRRAE ($\beta=0$):} A Variational RRAE without the KL divergence, to show its significance for generating meaningful samples.
\end{enumerate}

The errors and FID scores over all real-world datasets presented in the paper can be found in Table \ref{tbl:ab_results}.

\begin{table}[!ht]
  \caption{Quantitative comparison across datasets for the ablation study. Same metrics as in Table \ref{tbl:rw_results}.}
  \label{tbl:ab_results}
  \centering
  \resizebox{\textwidth}{!}{%
    \begin{tabular}{cccccccccc}
      \toprule
      {} & \multicolumn{3}{c}{MNIST} & \multicolumn{3}{c}{CIFAR-10} & \multicolumn{3}{c}{CelebA} \\
      \cmidrule(lr){2-4} \cmidrule(lr){5-7} \cmidrule(lr){8-10}
      Model & Inter. & Rand. & Rec. & Inter. & Rand. & Rec. & Inter. & Rand. & Rec. \\
      \midrule
      RRAE + VAE  & 20.07 & 71.96 & 44.81  & 160.23 & 158.36 & 29.35  & 12.71 & 17.50 & 32.80 \\
      VAE ($f=I$)   & 147.573 & 140.908  & 94.95   & 164.72 & 157.00 & 58.68  & 22.32  & 22.18  & 54.29  \\
      VRRAE ($\beta=0$) & 5.37  & 38.47 & 26.17 & \textbf{129.68}  & \textbf{129.89}  & \textbf{17.04} & 7.68  & 14.23  & 15.21  \\
      \textbf{VRRAE}      & \textbf{4.89}  & \textbf{38.77} & \textbf{26.00}  & \textbf{129.68} & \textbf{129.89} & \textbf{17.04}  & \textbf{7.06}  & \textbf{7.60}  & \textbf{15.03} \\
      \bottomrule
    \end{tabular}%
  }
\end{table}

A can be seen in both reconstruction errors and FIDs for interpolation/random generation, setting $f=I$ does not by itself bring any advantages when implemented on VAEs. However, it is necessary to get a good performance with VRRAEs. Further, the additional regularization caused by the KL divergence is necessary in most of the problems (except on CIFAR-10) to get a meaningful latent space (more details about the effect of the KL divergence can be found in \ref{ap:KL}).

\section{Conclusion}
All in all, we presented in this paper a novel architecture, Variational Rank Reduction Autoencoders. The main objective behind VRRAEs is to learn a probabilistic model to be able to generate new samples (similarly to a VAE), while benefiting from the regularization imposed by RRAEs. Our results show that by preserving the deterministic values as the mean values of the distribution (i.e., $f=I$), VRRAEs are better regularized than VAEs, and are more robust to posterior collapse. They also outperform significantly their deterministic version RRAEs. Empirically, the robustness to collapse was showcased on a synthetic dataset of small size, while the regularization effect was shown on three real-world datasets; MNIST, CelebA, and CIFAR-10, over which VRRAEs achieved better FID scores, over almost all generation/interpolation tasks compared to VAEs and RRAEs.

\begin{ack}
\noindent We thank SKF Magnetic Mechatronics for funding the research. We also thank the Google TPU Research cloud (TRC) program for giving us the resources needed to run all of the experiments.

\noindent This work was also supported by Ministerio de Asuntos Económicos y Transformación Digital, Gobierno de España (Grant No. TSI-100930-2023-1) and Ministerio de Ciencia, Innovación y Universidades (Grant No. PID2023-147373OB-I00).
\end{ack}

\newpage

\appendix
\section{Appendix}
\subsection{Naive RRAE + VAE}\label{ap:naive}
In this part, we present a naive combination of the concepts of an RRAE and a VAE, with both $f$ and $g$ being learnable linear maps. While this approach makes RRAEs variational, the sampled coefficients $\tilde{\alpha}$ are problematic for two main reasons:
\begin{enumerate}
    \item \underline{The singular values are not necessarily sorted anymore:}

First, note that any SVD coefficients $\alpha=SV^T$ can be written as,
\begin{equation}\label{eqn:form}
    \bar{\alpha} = \bar{S} \bar{V}^T = \begin{bmatrix}
        & & \bar{s}_1\bar{V}_1^T & & \\
        & & \vdots & & \\
        & &\bar{s}_{k^*}\bar{V}_{k^*}^T & &
    \end{bmatrix},
\end{equation}
where $s_i$ is the $i$-th singular value, and $V_i^T$ is the $i$-th right singular vector, with $V$ being orthonormal. Note that we can write the sampled coefficients $\tilde{\alpha}$ in row format as follows,
\begin{equation}\label{fig:s_not_sorted}
    \tilde{\alpha} = \begin{bmatrix}
        & & \tilde{\alpha}_1^T & & \\
        & & \vdots & & \\
        & & \tilde{\alpha}_{k^*}^T & &
    \end{bmatrix}=\begin{bmatrix}
        & & \displaystyle\|\tilde{\alpha}_1^T\|_2\,\,\frac{\tilde{\alpha}_1^T}{\|\tilde{\alpha}_1^T\|_2} & & \\
        & & \vdots & & \\
        & & \displaystyle\|\tilde{\alpha}_{k^*}^T\|_2\,\,\frac{\tilde{\alpha}_{k^*}^T}{\|\tilde{\alpha}_{k^*}^T\|_2} & &
    \end{bmatrix}=\begin{bmatrix}
        & & \tilde{s}_1\tilde{V}_1^T & & \\
        & & \vdots & & \\
        & & \tilde{s}_{k^*}\tilde{V}_{k^*}^T & &
    \end{bmatrix},
\end{equation}
which means $\tilde{s}_i = \|\tilde{\alpha}_i^T\|_2$, and $\tilde{V}_i^T=\frac{\tilde{\alpha}_{i}^T}{\|\tilde{\alpha}_{i}^T\|_2}$ are the sampled singular values/vectors respectively. However, by choosing $f$ and $g$ to be linear maps, we can not enforce that $\|\tilde{\alpha}_i^T\|_2\geq \|\tilde{\alpha}_j^T\|_2, \forall i<j$. Accordingly, nothing guarantees that $\tilde{s}_i\geq\tilde{s}_j, \forall i<j$, which means that the sampled singular values are not sorted. This complicates the training and reduces its stability. Further, it makes the proposal of an adaptive algorithm like the one proposed in \cite{mounayer2025rankreductionautoencoders} impossible.

\item \underline{The expected value of $\tilde{Y}$ is not an SVD of $Y$:}
Note that a crucial property of RRAEs is that the truncated latent space $\bar{Y}$ is found by truncating the SVD of the original latent space $Y$. By applying a function $f$ that changes the mean value of $\bar{\alpha}$, the expected value of the bottleneck is modified, and hence the SVD is only used to find the basis (not the coefficients). This halts the proof of convergence to a common basis provided by \cite{mounayer2025rankreductionautoencoders}. In practice, the training becomes less stable, and the convergence to a common basis to represent the whole data by a bottleneck is not guaranteed.
\end{enumerate}

Note that empirically, we showed that choosing $f$ to be a learnable linear map achieves worse results in the ablation study in Section \ref{sec:ablation}.

\subsection{The KL divergence in VRRAEs}\label{ap:KL}
We begin by deriving, with more details, the expression of the KL divergence for VRRAEs given in Equation \eqref{eq:final_eq_kl_vrrae}. To do so, we begin by noting the generic KL divergence, written as,
\begin{equation*}
    \sum_{j=1}^N\mathrm{KL}\left(q(\tilde{\alpha} \mid X_j) \,\|\, p(\tilde{\alpha})\right)\ = 0.5\,\,\text{sum}(\mathbf{1}_{k^*\times N}+\log(\bar{\alpha}_\sigma\odot\bar{\alpha}_\sigma)-\bar{\alpha}_\mu\odot \bar{\alpha}_\mu-\bar{\alpha}_\sigma\odot\bar{\alpha}_\sigma).
\end{equation*}
In what follows, we focus on the term $(\bar{\alpha}_\mu\odot \bar{\alpha}_\mu)$. Note that since $\bar{\alpha}_\mu=\bar{S}\bar{V}^T$, we can write,
\begin{equation*}
    \bar{\alpha}_\mu\odot \bar{\alpha}_\mu = \left(\bar{S}\bar{V}^T\right) \odot \left(\bar{S}\bar{V}^T\right)
\end{equation*}
Note that $\bar{S}$ is diagonal with $\bar{s}_i$ being the $i$-th diagonal entry. Hence, we can write,
\begin{equation*}
    \bar{\alpha}_\mu\odot \bar{\alpha}_\mu = \begin{bmatrix}
        & & \bar{s}_1\bar{V}_1^T & & \\
        & & \vdots & & \\
        & &\bar{s}_{k^*}\bar{V}_{k^*}^T & &
    \end{bmatrix} \odot \begin{bmatrix}
        & & \bar{s}_1\bar{V}_1^T & & \\
        & & \vdots & & \\
        & &\bar{s}_{k^*}\bar{V}_{k^*}^T & &
    \end{bmatrix} = \begin{bmatrix}
        & & \bar{s}_1^2\left(\bar{V}_1^T \odot \bar{V}_1^T\right) & & \\
        & & \vdots & & \\
        & &\bar{s}_{k^*}^2\left(\bar{V}_{k^*}^T \odot \bar{V}_{k^*}^T\right) & &
    \end{bmatrix},
\end{equation*}
Hence, the sum can be written as,
\begin{equation*}
    \text{sum}(\bar{\alpha}_\mu\odot \bar{\alpha}_\mu) = \text{sum}\left(\begin{bmatrix}
        & \text{sum}\left(\bar{s}_1^2\left(\bar{V}_1^T \odot \bar{V}_1^T\right)\right) & \\
        & \vdots & \\
        &\text{sum}\left(\bar{s}_{k^*}^2\left(\bar{V}_{k^*}^T \odot \bar{V}_{k^*}^T\right)\right) & 
    \end{bmatrix}\right)=\text{sum}\left(\begin{bmatrix}
        & \bar{s}_1^2\,\,\,\text{sum}\left(\bar{V}_1^T \odot \bar{V}_1^T\right) & \\
        & \vdots & \\
        & \bar{s}_{k^*}^2 \, \, \, \text{sum}\left(\bar{V}_{k^*}^T \odot \bar{V}_{k^*}^T\right) &
    \end{bmatrix}\right),
\end{equation*}
By noting that for any $i$, $\|\bar{V}_i\|_2^2 = \text{sum}(\bar{V}_i^T\odot\bar{V}_i^T) = 1$ (the right singular vectors are, we can say,
\begin{equation*}
    \text{sum}(\bar{\alpha}_\mu\odot \bar{\alpha}_\mu) = \text{sum}\left(\begin{bmatrix}
        \bar{s}_1^2\\
        \vdots \\
        \bar{s}_{k^*}^2 
    \end{bmatrix}\right)
    = \text{sum}\left(\left(\text{diag}\left(\bar{S}\right)\right)^2\right),
\end{equation*} 
which is the term written in Equation \ref{eq:final_eq_kl_vrrae}. Note that this term implicates that by regularizing the KL divergence, we enforce the singular values to be bounded.

In what follows, we investigate the effect of the KL divergence term on the training of VRRAEs. First, we plot the singular values of the training latent space once training is done on the MNIST for different values of $\beta$. The plots can be found in Figure \ref{fig:svs}.

\renewcommand{\thefigure}{A.2-\arabic{figure}}
\setcounter{figure}{0}
\renewcommand{\thetable}{A.3-\arabic{table}}
\setcounter{table}{0}
\begin{figure}[!ht]
    \centering
    \includegraphics[width=0.5\linewidth]{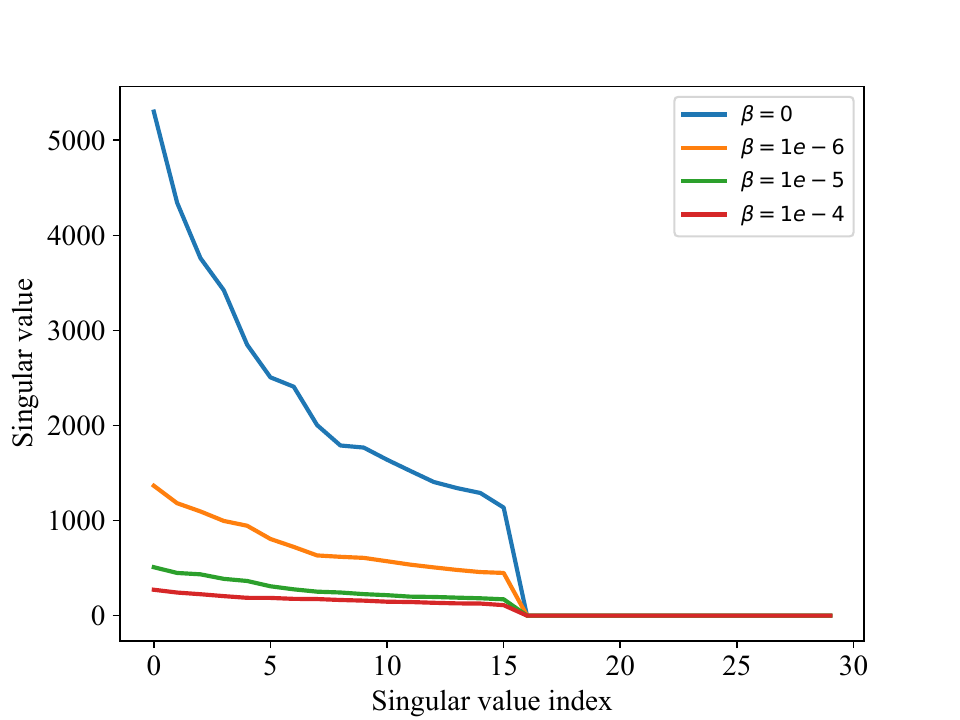}
    \caption{The singular values (by VRRAEs) of the training latent space on the MNIST dataset for different contributions of the KL divergence (i.e., different values of $\beta$). Note the bottleneck enforced with $k^*=16$.}
    \label{fig:svs}
\end{figure}

As expected, since the mean values of $\alpha$ depend on the singular values, the KL divergence enforces the singular values to become of smaller magnitude.

\begin{figure}[!ht]
    \centering
    \includegraphics[width=1\linewidth, trim=5cm 0 0 0, clip]{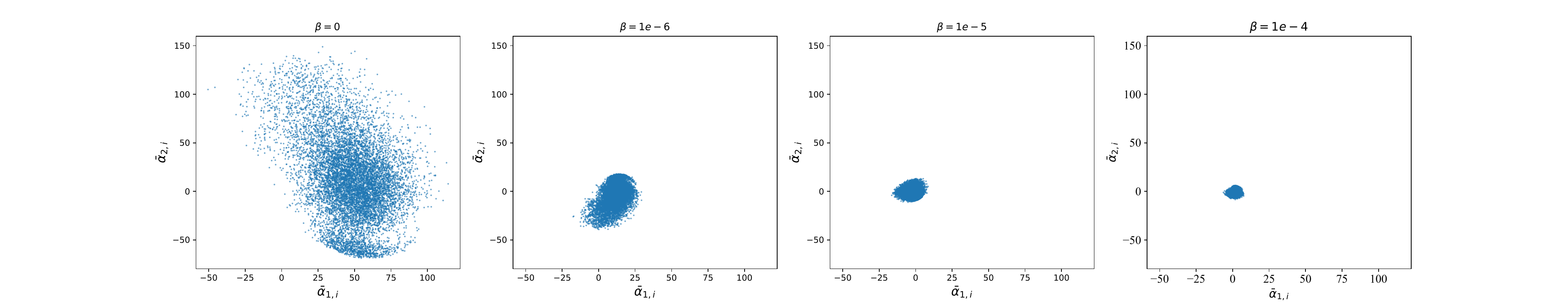}
    \caption{The mean of the second latent dimension against the mean of the first latent dimension for VRRAEs on the MNIST for different values of $\beta$.}
    \label{fig:kl}
\end{figure}

Further, we plot the mean value of the second latent dimension against the first one or the MNSIT dataset and it can be found in Figure \ref{fig:kl},

As expected, a higher value of $\beta$ enforces the latent space to be more meaningful by forcing the means to be closer to each others.

\subsection{Model Architecture and Training Parameters}\label{ap:model}
Throughout the paper, we used the same model architecture as the one proposed by \cite{mounayer2025rankreductionautoencoders}. In other words, our encoder had two convolution layers, and our decoder had four transpose convolution layers and an additional final convolution to match the size of the input. More details in \cite{mounayer2025rankreductionautoencoders}, Appendix C.

While all Autoencoder architectures share the same base architecture, the models have some differences. For instance, both RRAEs and VRRAEs have a latent size $L$, instead of reducing to the bottleneck $k^*$ as in VAEs and Diabolo AEs. The chosen values of $k^*$, and $L$, for each problem can be found in Table \ref{sample-table-nana}.

\begin{table}[!ht]
  \caption{Chosen values for $k^*$ (the bottleneck), and the latent size (for RRAEs, and VRRAEs) for each problem presented in the paper.}
  \label{sample-table-nana}
  \centering
  \begin{tabular}{ccc}
    \toprule
    Problem & $k^*$ & $L$\\
    \midrule
    Synthetic & 2 &   200   \\
    MNIST & 16 & 100  \\
    CIFAR-10 & 60 &  512  \\
    CelebA & 186 & 512 \\
    \bottomrule
\end{tabular}
\end{table}

On the other hand, note that both VAEs, and VRRAEs have an extra hyperparameter to tune $\beta$. The optimal value of $\beta$ changes from problem to another and could possibly be different for VAEs and VRRAEs. We trained both VAEs and VRRAEs for different values of $\beta$ and retained the best model. The FID for the images generated  for every value of $\beta$ can be found in Table \ref{fig:sample-table-nono}.

\begin{table}[!ht]
    \caption{Table documenting FID values for 20000 images generated by Random sampling over different datasets for different values of $\beta$.}
   \label{fig:sample-table-nono}
  \centering
  
  \begin{tabular}{cccccc}
        \toprule
\multicolumn{6}{c}{MNIST}                             \\
        \midrule                                                                                              \\
        \addlinespace[-0.2cm]
        $\beta$         & 0                      & 1e-6                     & 1e-4   & 1e-3 & 1e-2    \\
        \midrule
        VAE    &    40.8                         & 39.46                          & 35.21          & 30.63 & 40.1     \\
      VRRAE & 38.47                   & 38.77                  & 37.63 & 34.37 & 38.9 \\
      \midrule
        \multicolumn{6}{c}{CIFAR-10} \\
       \midrule
    $\beta$         & 0                      & 1e-5                     & 1e-4    \\
    \midrule
      VAE     & 135.55                            & 135.77                           & 153.11 \\
      VRRAE  & 129.89                          &  130.62                          & 145.29\\
    \midrule
        \multicolumn{6}{c}{CelebA} \\
       \midrule
    $\beta$         & 0                      & 1e-6                     & 1e-4 & 1e-3    \\
    \midrule
      VAE     & 14.65 & 9.65 & 13.31 & 17.42 \\
      VRRAE  & 7.68 & 7.6 & 13.03 & 17.9\\
        \bottomrule
    \end{tabular}
\end{table}

As can be seen in the Table, VRRAEs, even with $\beta=0$, can outperform VAEs due to the regularization strongly enforced by the SVD in the latent space.

For training, we choose different number of epochs, learning rates, and batch sizes for each dataset. These are documented in the Table \ref{tbl:params}.

\begin{table}[!ht]
  \caption{Chosen training parameters for all datasets in the paper.}
  \label{tbl:params}
  \centering
  \begin{tabular}{cccc}
    \toprule
    Problem & Epochs & Batch size & Learning rate\\
    \midrule
    Synthetic & 1280 &   64 & 1e-4   \\
    MNIST & 20 & 576 & 1e-3  \\
    CIFAR-10 & 5 &  64 & 1e-4  \\
    CelebA & 8 & 576 & 1e-3 \\
    \bottomrule
\end{tabular}
\end{table}
The adabeleif optimizer was used for all examples and all models. All codes were done in equinox (JAX) and were parallelized over 8 TPUs, hence why batch sizes are multiples of 8.

\renewcommand{\thetable}{A.4-\arabic{table}}
\setcounter{table}{0}

\subsection{Time complexity} \label{ap:time}
In this section, we study the added time complexity in VRRAEs. Note that the main difference between both VAEs and VRRAEs is the SVD computation in the latent space.  However, the latent matrix $Y$ is of size $(L \times \text{bs})$, where $L$ is the chosen latent
space dimension, and $\text{bs}$ is the batch size. Accordingly, the number of floating point operations (flops) for an SVD in a forward pass is:
\begin{equation*}
    \text{n}_\text{flops} = \mathcal{O} \, (L \times \text{bs} \times \min(\text{bs}, L))
\end{equation*}
On the other hand, note that the backward pass through the SVD is only a matrix multiplication (as shown in the Appendix of \cite{mounayer2025rankreductionautoencoders}). In addition, the matrices being multiplied during the backward passare all of dimensions smaller than  $\max(L, \text{bs})$. Consequently, the additional number of flops during backward propagation is around the following,
\begin{equation*}
    \text{n}_\text{back} = \mathcal{O} \, \max(L, \text{bs})^2
\end{equation*}
From the above, we can conclude that the additional time complexity of VRRAEs is similar to adding an additional layer of size $\max(L, \text{bs})$ to the network. While the overhead is not negligible, it can be considered small compared to the time complexity of the convolutional layers in the encoder/decoder. Empirically, both the required time for one forward/backward pass, as well as the total training time, can be found in Table \ref{tbl:time},

\begin{table}[!ht]
  \caption{Training times for one forward/backward pass ($t$), and the total training time ($T$) for both VAEs and VRRAEs on all real-world datasets.}
  \centering
  \label{tbl:time}
  \begin{tabular}{ccccccc}
    \toprule
    Model & $t_{\text{MNIST}}$ & $T_{\text{MNIST}}$ & $t_{\text{CIFAR}}$ & $T_{\text{CIFAR}}$ & $t_{\text{CelebA}}$ & $T_{\text{CelebA}}$ \\
    \midrule
    VAE	& 2.64 s &	92.4 mn	& 0.315 s	& 20.47 mn &	3.18 s &	116.6 mn \\
    VRRAE	& 2.79 s & 	97.7 mn &	0.319 s	& 20.74 mn	& 3.39 s	& 124.3 mn \\
    \bottomrule
\end{tabular}
\end{table}

\subsection{Effect of the batch size on the SVD} \label{ap:bs}
Since the SVD is computed on the latent space $Y$, which is of size $(L \times \text{bs})$, where $L$ is the latent space dimension, and $\text{bs}$ is the batch size, the batch size has an effect on the SVD computation. In what follows, we show empirically that the batch size does not have a significant effect on the performance of VRRAEs. To do so, we trained VRRAEs on the synthetic dataset (i.e. the 2D Gaussian curves) for different batch sizes, and we documented the test reconstruction error in Table \ref{tbl:bs_effect}.

\begin{table}[!ht]
  \caption{Test Reconstruction Errors (in \%) for different batch sizes on the synthetic dataset (i.e., 2D gaussians).}
  \centering
  \label{tbl:bs_effect}
  \begin{tabular}{ccccccc}
    \toprule
    Batch size & 8 & 16 & 32 & 64 \\
    \midrule
    Test error	& $10.03 \pm 8.96$	& $9.45 \pm 2.13$	& $8.47 \pm 3.6$	& $12.07 \pm 4.1$\\
    \bottomrule
\end{tabular}
\end{table}
Note that the mean values are similar. On the other hand, the standard deviation is higher for a very small batch size. However, this is expected behavior when training Neural Networks, so further study is needed to conclude wether the difference in standard deviation is due to the SVD or not when the batch size is as small as $8$.

\subsection{Comparison to other Autoencoders} \label{ap:other}
In this section, we compare VRRAEs to other Autoencoder architectures on both MNIST and CelebA datasets. The FID values for both interpolation and random generation are taken from \cite{mounayer2025rankreductionautoencoders}. The results can be found in Table \ref{tbl:other_ae}.

\begin{table}[!h]
  \caption{Quantitative comparison across datasets. The third column for each dataset represents the test reconstruction error, while the first two columns document the mean FID score (over 5 random seeds) for interpolation and random generation respectively. Standard deviations have been omitted since they're of small magnitudes.}
  \label{tbl:other_ae}
  \centering
  \begin{tabular}{ccccccc}
    \toprule
    {} & \multicolumn{3}{c}{MNIST} & \multicolumn{3}{c}{CelebA} \\
    \cmidrule(lr){2-4} \cmidrule(lr){5-7}
    Model & Inter. & Rand. & Rec. & Inter. & Rand. & Rec.\\
    \midrule
    Long    &    10.16                         & 86.5                          & 35.21          & 67.13 &   16.53 &   17.23     \\
      IRMAE & 8.09                   & 42.58                  & 23.2 & 43.62 & 15.79 & 15.47 \\
      Cont & 30.5                   & 90.5                  & 27.51  & 55.32  & 16.74 & 17.36 \\
      Sparse     & 6.1                          & 57.67                          & 61.32          & 87.2    & 16.91 & 16.57      \\
    AE     & 7.31  & 40.12 & 27.31 & 15.24 & 15.88 & 18.82 \\
    VAE    & 11.95  & \textbf{30.63} & 32.602  & 8.94 & 9.66 & 16.55 \\
    RRAE   & 6.68  & 45.46 & 27.90 & 13.27 & 13.70 & 17.52 \\
    \textbf{VRRAE}  & \textbf{5.89}  & 38.77 & \textbf{26.00}  & \textbf{7.06}  & \textbf{7.60}  & \textbf{15.03} \\
    \bottomrule
  \end{tabular}
\end{table}

\subsection{FID computation}\label{ap:FID}
Throughout the paper, we use the FID score to quantify how good generated/interpolated images are. The FID score was computed as follows,
\begin{enumerate}
    \item Loading an InceptionNet-v3, pretrained on ImagNet, from PyTroch.
    \item Removing the last layer, and replacing it with a layer of the right output size (e.g., $10$ for MNIST since we have 10 classes).
    \item Fine tune the InceptionNet by training it on the classification task of the corresponding dataset.
    \item Removing the last layer, the remaining part of the Network is a ``feature extractor''.
    \item The FID is computed between the feature vectors of the training data, and those of the generated images.
\end{enumerate}

\renewcommand{\thefigure}{A.5-\arabic{figure}}
\setcounter{table}{0}

\subsection{More Interpolations}\label{ap:cifar-samples}
\begin{figure}[!hb]
    \centering
    \includegraphics[width=0.9\linewidth, trim=0 0.5cm 0 0cm, clip]{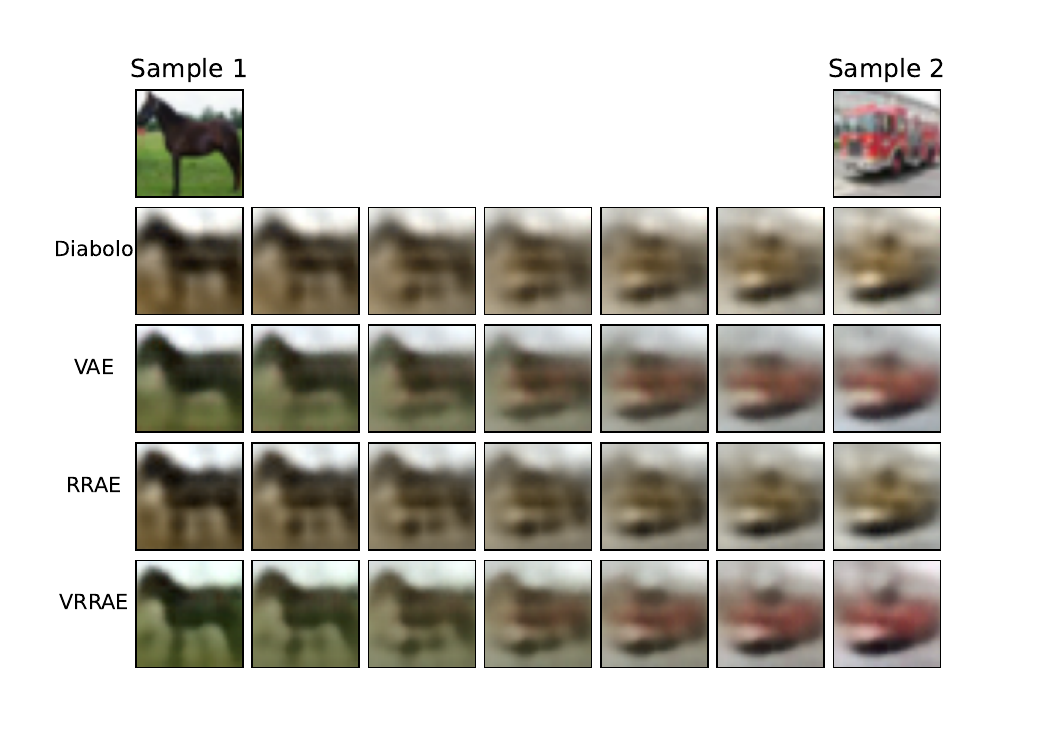}
    \caption{Example of linear interpolation in the latent space for the CIFAR-10 dataset.}
    \label{fig:cifar_interp}
\end{figure}

\begin{figure}[!ht]
    \centering
    \includesvg[width=0.9\linewidth]{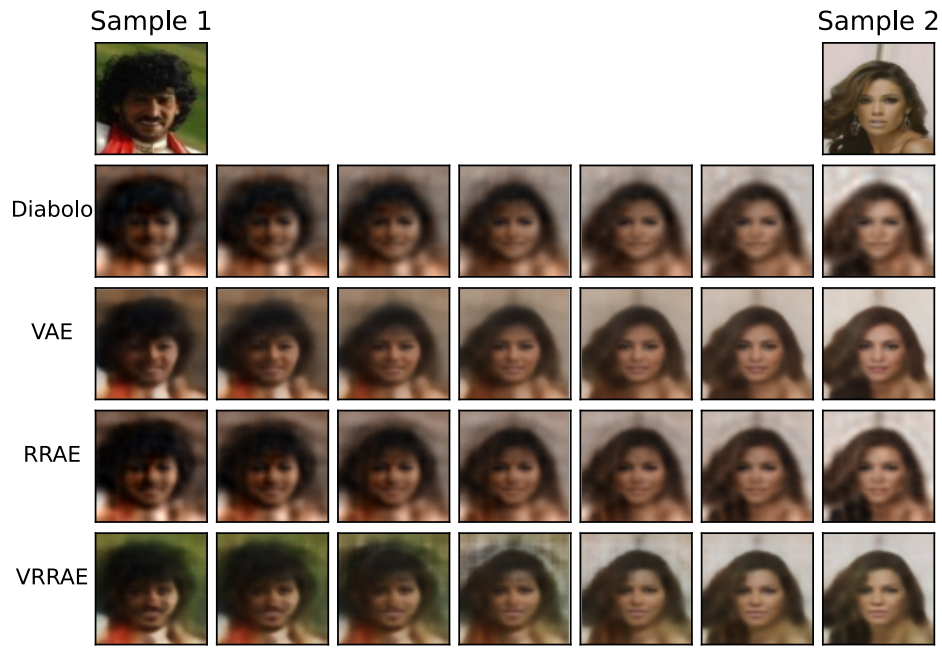}
    \caption{Example of linear interpolation in the latent space for the CelebA dataset (samples 159614 and 112203).}
    \label{fig:celeba_interp_ap_0}
\end{figure}

\begin{figure}[!ht]
    \centering
    \includesvg[width=0.9\linewidth]{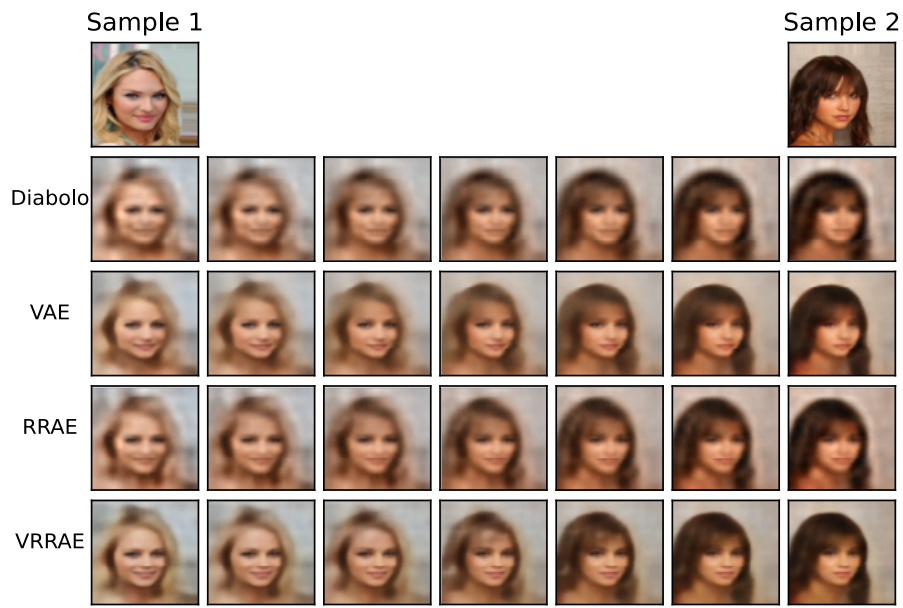}
    \caption{Example of linear interpolation in the latent space for the CelebA dataset (samples 49977 and 126035).}
    \label{fig:celeba_interp_ap_1}
\end{figure}

\begin{figure}[!ht]
    \centering
    \includegraphics[width=0.9\linewidth]{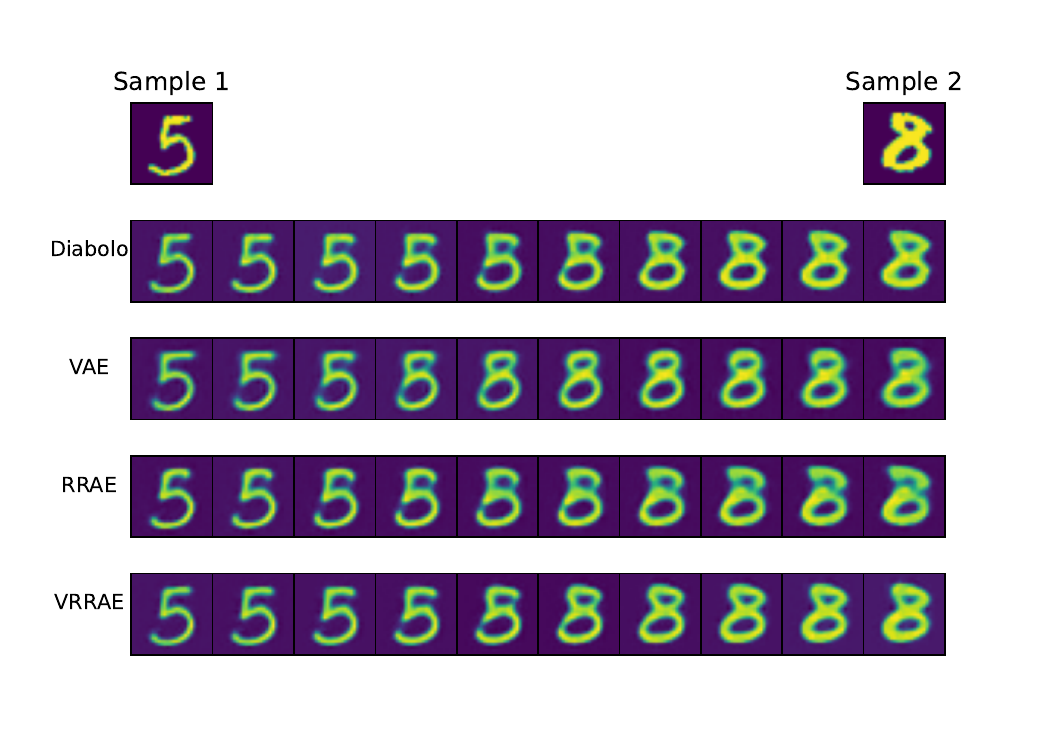}
    \caption{Example of linear interpolation in the latent space for the MNIST dataset.}
    \label{fig:mnist_interp}
\end{figure}

Since CIFAR-10 images are of low resolution ($28 \times 28$), and the autoencoders compress these images, the resulting photos are of low quality. An example of interpolated images of the CIFAR-10 dataset can be seen in Figure \ref{fig:cifar_interp}, more images can be found in Figures \ref{fig:celeba_interp_ap_0}, \ref {fig:celeba_interp_ap_1}, and \ref{fig:mnist_interp}, where VRRAEs have the sharpest reconstruction and the best interpolation compared to both VAEs and RRAEs.
Since CIFAR-10 images are of low resolution ($28 \times 28$), and the autoencoders compress these images, the resulting photos are of low quality. An example of interpolated images of the CIFAR-10 dataset can be seen in Figure \ref{fig:cifar_interp}, more images can be found in Figures \ref{fig:celeba_interp_ap_0}, \ref {fig:celeba_interp_ap_1}, and \ref{fig:mnist_interp}, where VRRAEs have the sharpest reconstruction and the best interpolation compared to both VAEs and RRAEs.

\begin{figure}[!ht]
    \centering
    \includesvg[width=0.8\linewidth]{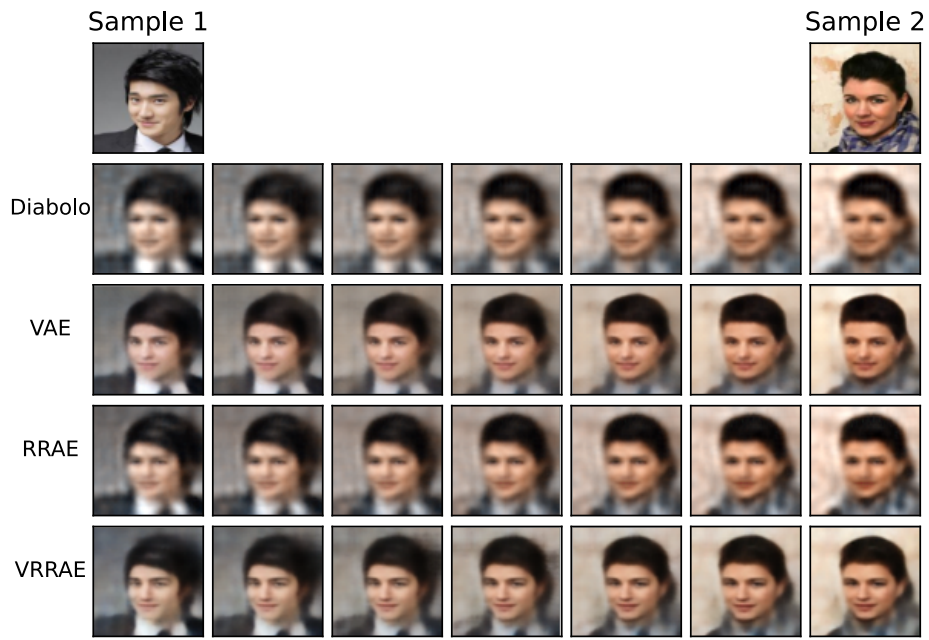}
    \caption{Example of linear interpolation in the latent space for the CelebA dataset (samples 96458 and 30835).}
    \label{fig:celeba_interp_ap_2}
\end{figure}

\clearpage
\bibliographystyle{elsarticle-num} 
\bibliography{biblio}

\begin{thebibliography}{10}
\expandafter\ifx\csname url\endcsname\relax
  \def\url#1{\texttt{#1}}\fi
\expandafter\ifx\csname urlprefix\endcsname\relax\def\urlprefix{URL }\fi
\expandafter\ifx\csname href\endcsname\relax
  \def\href#1#2{#2} \def\path#1{#1}\fi

\bibitem{jax}
J.~Bradbury, R.~Frostig, P.~Hawkins, M.~J. Johnson, C.~Leary, D.~Maclaurin,
  G.~Necula, A.~Paszke, J.~VanderPlas, S.~W. Milne, Q.~Zhang,
  \href{https://github.com/google/jax}{Jax: composable transformations of
  python + numpy programs} (2018).
\newline\urlprefix\url{https://github.com/google/jax}

\bibitem{eqx}
P.~Kidger, C.~Garcia, \href{https://github.com/patrick-kidger/equinox}{Equinox:
  neural networks in jax via callable pytrees and filtered transformations},
  differentiable Programming Workshop at NeurIPS 2021 (2021).
\newline\urlprefix\url{https://github.com/patrick-kidger/equinox}

\bibitem{planck1900zur}
M.~Planck, Zur theorie des gesetzes der energieverteilung im normalspektrum,
  Verhandlungen der Deutschen Physikalischen Gesellschaft 2 (1900) 237--245,
  translated as "On the Theory of the Energy Distribution Law in the Normal
  Spectrum".

\bibitem{wetterich2020probabilistic}
C.~Wetterich, \href{https://arxiv.org/abs/2007.00895}{The probabilistic world},
  arXiv preprint arXiv:2007.00895 (2020).
\newline\urlprefix\url{https://arxiv.org/abs/2007.00895}

\bibitem{reynolds2009gaussian}
D.~A. Reynolds, et~al., Gaussian mixture models., Encyclopedia of biometrics
  741~(659-663) (2009) 3.

\bibitem{mcnicholas2008parsimonious}
P.~D. McNicholas, T.~B. Murphy, Parsimonious gaussian mixture models,
  Statistics and Computing 18 (2008) 285--296.

\bibitem{pearl1988probabilistic}
J.~Pearl, Probabilistic Reasoning in Intelligent Systems: Networks of Plausible
  Inference, Morgan Kaufmann, San Mateo, CA, 1988.

\bibitem{govan2018bayesiannetwork}
P.~B. Govan, Bayesiannetwork: Interactive bayesian network modeling and
  analysis, Journal of Open Source Software 3~(21) (2018) 425.
\newblock \href {https://doi.org/10.21105/joss.00425}
  {\path{doi:10.21105/joss.00425}}.

\bibitem{govan2023bayesiannetwork}
P.~Govan,
  \href{https://CRAN.R-project.org/package=BayesianNetwork}{BayesianNetwork:
  Bayesian Network Modeling and Analysis}, r package version 0.3 (2023).
\newblock \href {https://doi.org/10.32614/CRAN.package.BayesianNetwork}
  {\path{doi:10.32614/CRAN.package.BayesianNetwork}}.
\newline\urlprefix\url{https://CRAN.R-project.org/package=BayesianNetwork}

\bibitem{rabiner1986introduction}
L.~R. Rabiner, A tutorial on hidden markov models and selected applications in
  speech recognition, Proceedings of the IEEE 77~(2) (1986) 257--286.

\bibitem{harte2021hiddenmarkov}
D.~Harte, \href{https://www.statsresearch.co.nz/dsh/sslib/}{HiddenMarkov:
  Hidden Markov Models}, Statistics Research Associates, Wellington, r package
  version 1.8-13 (2021).
\newline\urlprefix\url{https://www.statsresearch.co.nz/dsh/sslib/}

\bibitem{bourlard1994hidden}
H.~A. Bourlard, N.~Morgan, Hidden markov models, in: Connectionist Speech
  Recognition, Springer, Boston, MA, 1994, pp. 247--296.

\bibitem{gelman1996markov}
A.~Gelman, D.~B. Rubin, Markov chain monte carlo methods in biostatistics,
  Statistical Methods in Medical Research 5~(4) (1996) 399--415.

\bibitem{tierney1994markov}
L.~Tierney, Markov chains for exploring posterior distributions, Annals of
  Statistics 22~(4) (1994) 1701--1762.

\bibitem{kingma2013autoencoding}
D.~P. Kingma, M.~Welling, \href{https://arxiv.org/abs/1312.6114}{Auto-encoding
  variational bayes}, arXiv preprint arXiv:1312.6114Accessed: 2025-04-24
  (2013).
\newline\urlprefix\url{https://arxiv.org/abs/1312.6114}

\bibitem{doersch2021tutorialvariationalautoencoders}
C.~Doersch, \href{https://arxiv.org/abs/1606.05908}{Tutorial on variational
  autoencoders} (2021).
\newblock \href {http://arxiv.org/abs/1606.05908} {\path{arXiv:1606.05908}}.
\newline\urlprefix\url{https://arxiv.org/abs/1606.05908}

\bibitem{DBLP:journals/corr/abs-1906-02691}
D.~P. Kingma, M.~Welling, \href{http://arxiv.org/abs/1906.02691}{An
  introduction to variational autoencoders}, CoRR abs/1906.02691 (2019).
\newblock \href {http://arxiv.org/abs/1906.02691} {\path{arXiv:1906.02691}}.
\newline\urlprefix\url{http://arxiv.org/abs/1906.02691}

\bibitem{laptev2021generative}
V.~V. Laptev, O.~M. Gerget, N.~A. Markova, Generative models based on vae and
  gan for new medical data synthesis, Society 5.0: Cyberspace for advanced
  human-centered society (2021) 217--226.

\bibitem{lovric2021should}
M.~Lovri{\'c}, T.~{\DJ}uri{\v{c}}i{\'c}, H.~T. Tran, H.~Hussain, E.~Laci{\'c},
  M.~A. Rasmussen, R.~Kern, Should we embed in chemistry? a comparison of
  unsupervised transfer learning with pca, umap, and vae on molecular
  fingerprints, Pharmaceuticals 14~(8) (2021) 758.

\bibitem{yan2020chiller}
K.~Yan, J.~Su, J.~Huang, Y.~Mo, Chiller fault diagnosis based on vae-enabled
  generative adversarial networks, IEEE Transactions on Automation Science and
  Engineering 19~(1) (2020) 387--395.

\bibitem{regenwetter2022deep}
L.~Regenwetter, A.~H. Nobari, F.~Ahmed, Deep generative models in engineering
  design: A review, Journal of Mechanical Design 144~(7) (2022) 071704.

\bibitem{gao2020zero}
R.~Gao, X.~Hou, J.~Qin, J.~Chen, L.~Liu, F.~Zhu, Z.~Zhang, L.~Shao,
  Zero-vae-gan: Generating unseen features for generalized and transductive
  zero-shot learning, IEEE Transactions on Image Processing 29 (2020)
  3665--3680.

\bibitem{niu2020lstm}
Z.~Niu, K.~Yu, X.~Wu, Lstm-based vae-gan for time-series anomaly detection,
  Sensors 20~(13) (2020) 3738.

\bibitem{wu2020vector}
H.~Wu, M.~Flierl, Vector quantization-based regularization for autoencoders,
  in: Proceedings of the AAAI Conference on Artificial Intelligence, 2020, pp.
  6380--6387.

\bibitem{hadjeres2017glsr}
G.~Hadjeres, F.~Nielsen, F.~Pachet, Glsr-vae: Geodesic latent space
  regularization for variational autoencoder architectures, in: 2017 IEEE
  symposium series on computational intelligence (SSCI), IEEE, 2017, pp. 1--7.

\bibitem{tolstikhin2019wassersteinautoencoders}
I.~Tolstikhin, O.~Bousquet, S.~Gelly, B.~Schoelkopf,
  \href{https://arxiv.org/abs/1711.01558}{Wasserstein auto-encoders} (2019).
\newblock \href {http://arxiv.org/abs/1711.01558} {\path{arXiv:1711.01558}}.
\newline\urlprefix\url{https://arxiv.org/abs/1711.01558}

\bibitem{bredell2023explicitlyminimizingblurerror}
G.~Bredell, K.~Flouris, K.~Chaitanya, E.~Erdil, E.~Konukoglu,
  \href{https://arxiv.org/abs/2304.05939}{Explicitly minimizing the blur error
  of variational autoencoders} (2023).
\newblock \href {http://arxiv.org/abs/2304.05939} {\path{arXiv:2304.05939}}.
\newline\urlprefix\url{https://arxiv.org/abs/2304.05939}

\bibitem{dalal2024shorttimefouriertransformdeblurring}
V.~Dalal, \href{https://arxiv.org/abs/2401.03166}{Short-time fourier transform
  for deblurring variational autoencoders} (2024).
\newblock \href {http://arxiv.org/abs/2401.03166} {\path{arXiv:2401.03166}}.
\newline\urlprefix\url{https://arxiv.org/abs/2401.03166}

\bibitem{khan2018adversarialtrainingvariationalautoencoders}
S.~H. Khan, M.~Hayat, N.~Barnes,
  \href{https://arxiv.org/abs/1804.10323}{Adversarial training of variational
  auto-encoders for high fidelity image generation} (2018).
\newblock \href {http://arxiv.org/abs/1804.10323} {\path{arXiv:1804.10323}}.
\newline\urlprefix\url{https://arxiv.org/abs/1804.10323}

\bibitem{he2019lagginginferencenetworksposterior}
J.~He, D.~Spokoyny, G.~Neubig, T.~Berg-Kirkpatrick,
  \href{https://arxiv.org/abs/1901.05534}{Lagging inference networks and
  posterior collapse in variational autoencoders} (2019).
\newblock \href {http://arxiv.org/abs/1901.05534} {\path{arXiv:1901.05534}}.
\newline\urlprefix\url{https://arxiv.org/abs/1901.05534}

\bibitem{havrylov2020preventingposteriorcollapselevenshtein}
S.~Havrylov, I.~Titov, \href{https://arxiv.org/abs/2004.14758}{Preventing
  posterior collapse with levenshtein variational autoencoder} (2020).
\newblock \href {http://arxiv.org/abs/2004.14758} {\path{arXiv:2004.14758}}.
\newline\urlprefix\url{https://arxiv.org/abs/2004.14758}

\bibitem{dang2024vanillavariationalautoencodersdetecting}
H.~Dang, T.~Tran, T.~Nguyen, N.~Ho,
  \href{https://arxiv.org/abs/2306.05023}{Beyond vanilla variational
  autoencoders: Detecting posterior collapse in conditional and hierarchical
  variational autoencoders} (2024).
\newblock \href {http://arxiv.org/abs/2306.05023} {\path{arXiv:2306.05023}}.
\newline\urlprefix\url{https://arxiv.org/abs/2306.05023}

\bibitem{mounayer2025rankreductionautoencoders}
J.~Mounayer, S.~Rodriguez, C.~Ghnatios, C.~Farhat, F.~Chinesta,
  \href{https://arxiv.org/abs/2405.13980}{Rank reduction autoencoders} (2025).
\newblock \href {http://arxiv.org/abs/2405.13980} {\path{arXiv:2405.13980}}.
\newline\urlprefix\url{https://arxiv.org/abs/2405.13980}

\end{thebibliography}
\end{document}